\documentclass[10pt,twocolumn,letterpaper]{article}

\usepackage{wacv}              

\usepackage{graphicx}
\usepackage{amsmath}
\usepackage{amssymb}
\usepackage{booktabs}
\usepackage{multirow}

\usepackage[pagebackref,breaklinks,colorlinks]{hyperref}
\usepackage[accsupp]{axessibility}

\usepackage[capitalize]{cleveref}
\crefname{section}{Sec.}{Secs.}
\Crefname{section}{Section}{Sections}
\Crefname{table}{Table}{Tables}
\crefname{table}{Tab.}{Tabs.}

\def\wacvPaperID{387} 
\def\confName{WACV}
\def\confYear{2025}

\begin{document}

\title{Dropout the High-rate Downsampling: \\A Novel Design Paradigm for UHD Image Restoration}

\author{
    Chen Wu$^1$ \qquad
    Ling Wang$^2$ \qquad
    Long Peng$^1$ \qquad
    Dianjie Lu$^3$ \qquad
    Zhuoran Zheng$^4$ \thanks{Corresponding author}\\
    $^1$University of Science and Technology of China \\
    $^2$The Hong Kong University of Science and Technology (Guangzhou) \\
    $^3$Shandong Normal University \\
    $^4$Sun Yat-sen University \\
    {\tt\small wuchen5X@mail.ustc.edu.cn}
}
\maketitle

\begin{abstract}
With the popularization of high-end mobile devices, Ultra-high-definition (UHD) images have become ubiquitous in our lives. The restoration of UHD images is a highly challenging problem due to the exaggerated pixel count, which often leads to memory overflow during processing. Existing methods either downsample UHD images at a high rate before processing or split them into multiple patches for separate processing. However, high-rate downsampling leads to significant information loss, while patch-based approaches inevitably introduce boundary artifacts. In this paper, we propose a novel design paradigm to solve the UHD image restoration problem, called D2Net. D2Net enables direct full-resolution inference on UHD images without the need for high-rate downsampling or dividing the images into several patches. Specifically, we ingeniously utilize the characteristics of the frequency domain to establish long-range dependencies of features. Taking into account the richer local patterns in UHD images, we also design a multi-scale convolutional group to capture local features. Additionally, during the decoding stage, we dynamically incorporate features from the encoding stage to reduce the flow of irrelevant information. Extensive experiments on three UHD image restoration tasks, including low-light image enhancement, image dehazing, and image deblurring, show that our model achieves better quantitative and qualitative results than state-of-the-art methods. 
\end{abstract}

\section{Introduction}
With the widespread adoption of high-end mobile devices, such as smartphones, high-resolution images have become almost mainstream, capable of capturing more detailed visual content. However, various factors often lead to poor image quality during the imaging process, such as insufficient lighting conditions and unfavorable weather conditions.
Image restoration constitutes a significant and demanding undertaking in the field of computer vision, aiming to recover degraded images into clear, realistic, and clean images. Although a series of promising methods have emerged for the task of image restoration, achieving satisfactory results, especially the learning-based methods~\cite{ren2016single,Uformer,Restormer,Retinexformer,yang2023dual,li2019underwater,ren2018gated,ren2019low,cui2024revitalizing,cui2023dual,wang2023decoupling,yan2023textual,xiao2024towards,xiao2023online,wu2024seesr}, these methods face challenges when it comes to transferring them to Ultra-high-definition (UHD) images. The high pixel count of UHD images (\ie, 3840 $\times$ 2160) often leads to memory overflow issues during the processing of these methods.

\begin{figure}[t!]
    \centering
    \includegraphics[width=\linewidth]{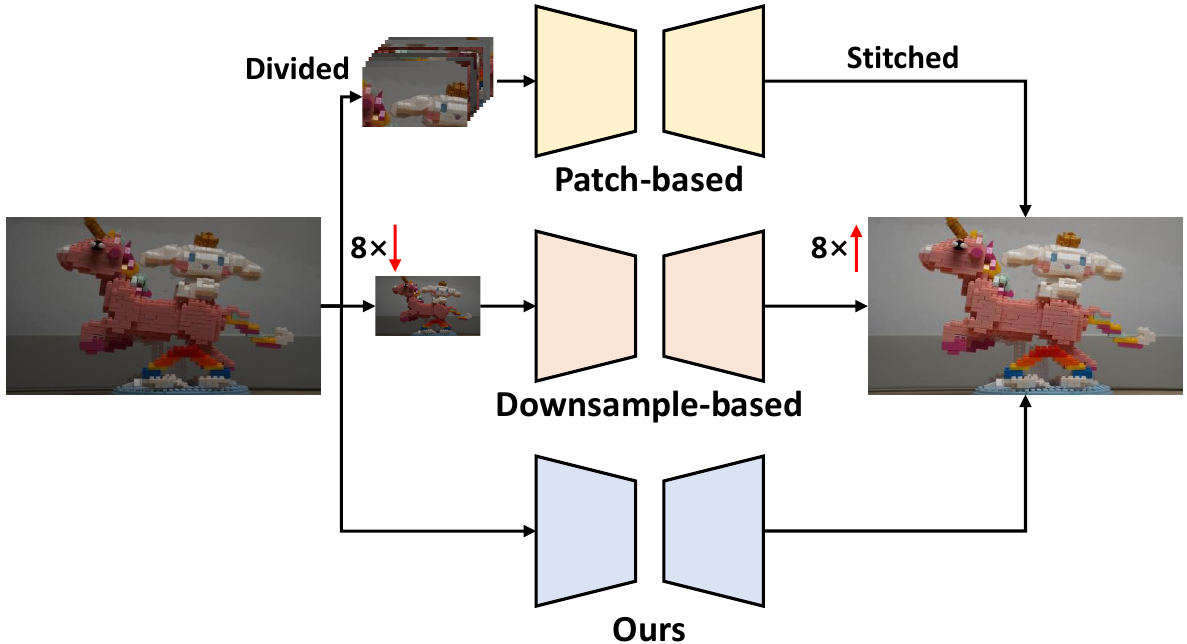}
    \caption{Comparison between previous methods and our proposed method for UHD image restoration. Due to hardware limitations, previous methods had to resort to patch-based or downsample-based approaches to enable consumer-grade GPUs to process UHD images. However, patch-based methods~\cite{LLFormer} introduce boundary artifacts during subsequent stitching processes, while downsample-based methods~\cite{UHDFour,NSEN} result in significant information loss, both of which can affect the quality of image enhancement. In contrast, our D2Net allows for direct full-resolution inference on UHD images.}
    \label{fig:diff}
\end{figure}

To tackle this issue, there are currently two primary processing methods: patch-based methods~\cite{LLFormer} and downsample-based methods~\cite{NSEN,UHDFour,MixNet}, as illustrated in Fig.~\ref{fig:diff}. Patch-based methods refer to the approach of dividing UHD images into non-overlapping patches during processing. These patches are individually processed and then stitched to form the final restored image. However, this method inevitably introduces boundary artifacts during the stitching process. Additionally, the process of dividing and stitching patches incurs additional time overhead. Downsample-based methods involve high-rate downsampling of UHD images before processing. After completing the image restoration steps, the images are then upsampled to restore their original resolution. Undoubtedly, high-rate downsampling leads to a significant loss of valuable information, which impacts the quality of the final enhanced image.

In this paper, we design a novel framework to solve the UHD image restoration problem called D2Net, which allows for direct full-resolution inference of UHD images. This represents a new design paradigm that completely discards the conventional approaches to UHD image restoration (patch-based and downsample-based). Specifically, we delve deeper into the characteristics of the Fourier domain, as previous works primarily focus on the Fourier transform's capability to some extent in disentangling image degradation and content components. We find that when an image is transformed from the spatial domain to the frequency domain, the features in the frequency domain naturally exhibit interactions. In light of this, we design the Fourier-based Global Feature Extraction (FGFE) Module to capture long-range dependencies of features. In addition, UHD images, due to their high resolution, contain richer and more complex patterns. Therefore, we design the Multi-scale Local Feature Extraction (MLFE) Module to capture the multi-scale local features of the UHD images. It is worth noting that the convolution kernel sizes in the MLFE module are relatively large. This is because, considering the high-resolution characteristics of UHD images, the commonly used $1\times1$ convolution or $3\times3$ convolution are often unable to capture high-quality local features. Moreover, we also design an Adaptive Feature Modulation Module (AFMM) that dynamically adjusts the fusion of encoded and decoded features. Unlike the simple stacking of features used in previous approaches, this method effectively suppresses the flow of irrelevant information, leading to enhanced performance. Extensive experiments across three datasets demonstrate that D2Net surpasses prior arts in terms of restoration quality and generalization ability.

In summary, our main contributions are three-fold: 
\begin{itemize}
    \item To the best of our knowledge, we propose the first UHD restoration method that enables direct full-resolution inference on consumer-grade GPUs, without the need for high-rate downsampling or dividing the images into several patches.
    
    \item We design a novel UHD restoration framework named D2Net. D2Net leverages the FGFE module to achieve long-range modeling and utilizes the MLFE module to capture multi-scale local features. In addition, we design the AFMM to dynamically adjust the fusion of encoded and decoded features.

    \item We conduct comprehensive quantitative and qualitative evaluations of our method, which demonstrate that D2Net achieves state-of-the-art results across multiple UHD image restoration tasks.
\end{itemize}

\section{Related Work}
\subsection{UHD Image Restoration}
In recent years, data-driven CNN~\cite{ren2016single,yang2023dual,li2019underwater,ren2018gated,ren2019low,cui2024revitalizing,wang2023decoupling} or Transformer~\cite{Restormer,Retinexformer,Uformer,cui2023dual,chen2023learning,yan2023textual,liang2021swinir,chen2022cross,xiao2022stochastic} architectures have been shown to out perform conventional restoration approaches. However, these methods often focus on processing low-resolution images ($400 \times 600$), and when faced with high-resolution images ($3840 \times 2160$), they often encounter problems such as memory overflow. 

Learning-based UHD image restoration has received some attention in recent years. Zheng~\etal~\cite{UHDHaze} incorporate bilateral learning into deep learning by using a network to learn local affine coefficients of the bilateral grid from low-resolution images and apply them to high-resolution images. Subsequently, several methods based on deep bilateral learning have emerged~\cite{UHDHDR,zhou20244k}. LLFormer~\cite{LLFormer} divides the UHD image into multiple non-overlapping patches and processes them individually, treating the UHD image as multiple low-resolution images. Li~\etal~\cite{UHDFour} discover that a high-resolution image and its corresponding low-resolution version exhibit similar amplitude patterns. Based on this observation, they propose UHDFour which restores the low-light UHD images in the frequency domain. Similarly, the UHD-SFNet~\cite{uhdUnderwater} also attempts to enhance underwater UHD images in the frequency domain. UHDformer~\cite{UHDformer} designs two branches to handle UHD images, placing the global feature modeling on the low-resolution branch to avoid memory overflow, significantly reducing the computational complexity of the model.
However, these methods introduce boundary artifacts and incur additional time overhead when patch-based strategies are employed, while downsample-based strategies result in a significant loss of useful information. These factors adversely affect the quality of the restored images. Our method performs full-resolution inference directly on UHD images, thereby avoiding the problems mentioned above.

\begin{figure*}[t!]
    \centering
    \includegraphics[width=\linewidth]{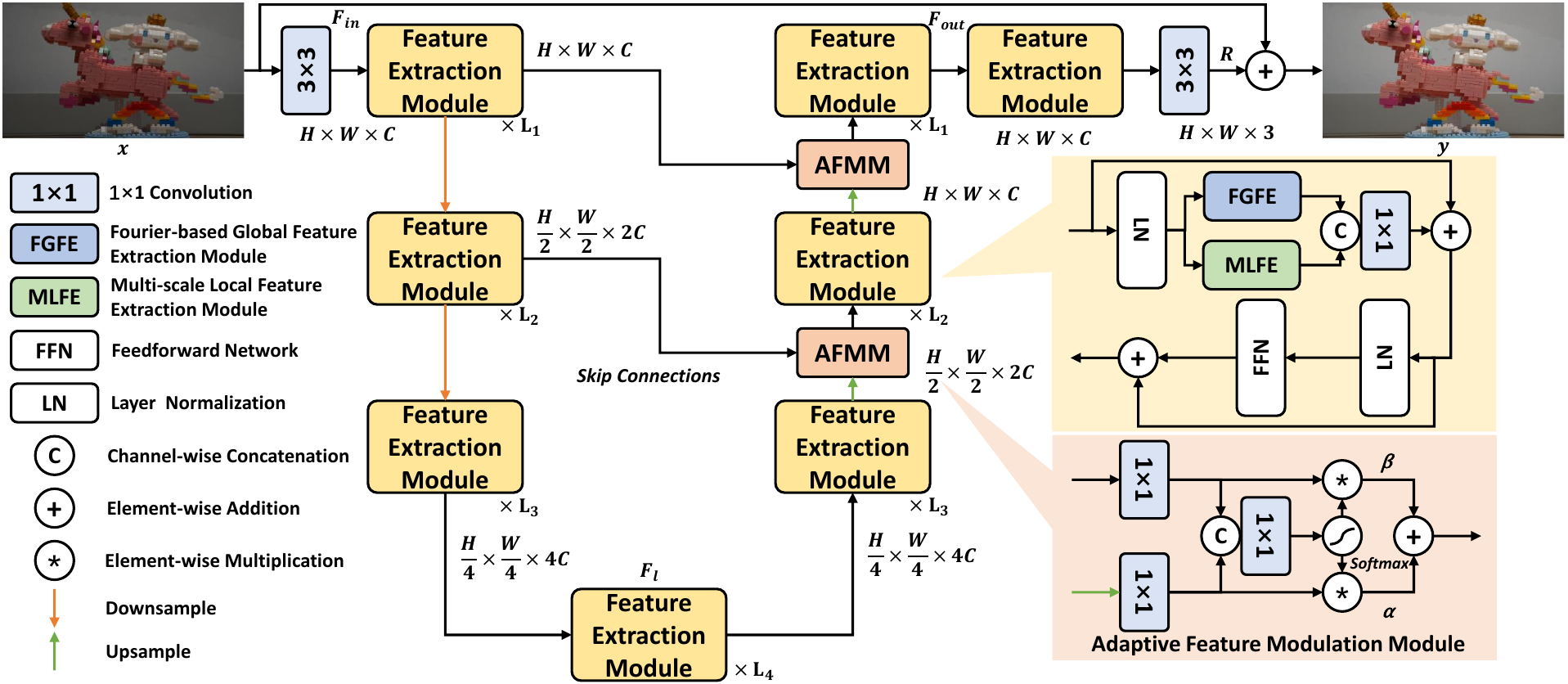}
    \vspace{-18pt}
    \caption{An overview of the network architecture of our D2Net. The degraded image is forwarded to a 4-level hierarchical encoder-decoder Unet-like structure to get a normal-light image. D2Net primarily consists of several Feature Extraction Modules (FEMs) and an Adaptive Feature Modulation Modules (AFMMs). FEM is responsible for feature extraction through global and local feature modeling. AFMM dynamically adjusts the fusion of encoded and decoded features to suppress the flow of irrelevant information.}
    \label{fig:framework}
\end{figure*}

\subsection{Image Restoration in Frequency Domain}
Learning with frequency domain has become increasingly common in low-level tasks. DWSR~\cite{guo2017deep} is trained in the wavelet domain with four input and output channels respectively for super-resolution tasks. Compared to learning with the wavelet domain, learning with the Fourier domain is more commonly used in image restoration tasks. Fourier transform has been widely employed for tasks such as image de-raining~\cite{guo2022exploring}, image dehazing~\cite{yu2022frequency}, low-light image enhancement~\cite{wang2023fourllie,UHDFour}, image deblurring~\cite{fftformer,mao2023intriguing}, image artifact removal~\cite{yang2023dual} and image warping~\cite{xiao2024towards}. These methods are mostly based on the observation that tshe Fourier transform can effectively decouple degraded information from image content. For instance, in low-light image enhancement tasks, a significant portion of illumination information is captured in the amplitude components, while the phase component contains most of the structure representation. In contrast to these methods, we focus on the excellent global modeling capability of the Fourier domain. When an image undergoes Fourier transformation, the features in the frequency domain inherently exhibit interactions. This greatly encourage us to try using Fourier transform to capture the long-range dependencies between features.

\section{Method}
In this section, we first take a brief review of the Fourier transform, and then we provide the overall pipeline of our method and further details on the critical components of our approach below.

\subsection{Preliminary}
The Fourier transform is an extensively employed methodology for analyzing the frequency characteristics of an image~\cite{frigo1998fftw}. In the case of images possessing multiple color channels, the Fourier transform is individually applied to each channel. Given an image $\boldsymbol{x}\in\mathbb{R}^{\mathrm{H\times W\times C}}$, the Fourier transform $\mathcal{F}$ converts it from the spatial domain to Fourier domain as the complex component $\mathcal{F}(\boldsymbol{x})$, which is expressed as: 
\begin{equation}
    \mathcal{F}(\boldsymbol{x})(\boldsymbol{u},\boldsymbol{v})=\frac{1}{\sqrt{\mathrm{HW}}}\sum_{\boldsymbol{h}=0}^{\mathrm{H}-1}\sum_{\boldsymbol{w}=0}^{\mathrm{W}-1}\boldsymbol{x}(\boldsymbol{h},\boldsymbol{w})e^{-j2\pi(\frac{\boldsymbol{h}}{\mathrm{H}}u+\frac{\boldsymbol{w}}{\mathrm{W}}\boldsymbol{v})},
    \label{eq:fourier}
\end{equation}
where $\boldsymbol{u}$ and $\boldsymbol{v}$ and the coordinates of the Fourier domain. $\mathcal{F}^{-1}(\cdot)$ defines the inverse Fourier transform. The amplitude component $\mathcal{A}(\boldsymbol{x})(\boldsymbol{u},\boldsymbol{v})$ and phase component $\mathcal{P}(\boldsymbol{x})(\boldsymbol{u},\boldsymbol{v})$ are expressed as:
\begin{equation}
    \begin{aligned}&\mathcal{A}(\boldsymbol{x})(\boldsymbol{u},\boldsymbol{v}))=\sqrt{\mathcal{R}^{2}(\boldsymbol{x})(\boldsymbol{u},\boldsymbol{v}))+\mathcal{I}^{2}(\boldsymbol{x})(\boldsymbol{u},\boldsymbol{v}))},\\&\mathcal{P}(\boldsymbol{x})(\boldsymbol{u},\boldsymbol{v}))=\arctan[\frac{\mathcal{I}(\boldsymbol{x})(\boldsymbol{u},\boldsymbol{v}))}{\mathcal{R}(\boldsymbol{x})(\boldsymbol{u},\boldsymbol{v}))}],\end{aligned}
    \label{eq:fourier_2}
\end{equation}
where $\mathcal{R}(\boldsymbol{x})(\boldsymbol{u,v})$ and $\mathcal{I}(\boldsymbol{x} (u,\boldsymbol{v})$ represent the real and imaginary parts respectively.

By examining Eqs.~\ref{eq:fourier} and \ref{eq:fourier_2}, it becomes evident that once an image is mapped to the Fourier domain, inherent correlations exist among individual pixels. This motivated us to develop a method for modeling long-range feature dependencies in the Fourier domain.

\subsection{Overview}
An overview of D2Net is shown in Fig.~\ref{fig:framework}. For a given degraded image $\boldsymbol{x}\in\mathbb{R}^{\mathrm{H\times W\times 3}}$, we first map it to the feature space through a $3\times3$ convolution to obtain the deep features $\boldsymbol{F_{in}}\in\mathbb{R}^{\mathrm{H\times W\times C}}$; where $\mathrm{H\times W}$ is the spatial resolution and $\mathrm{C}$ denotes the channel. Subsequently, deep features $\boldsymbol{F_{in}}$ will pass through an encoder to obtain latent features $\boldsymbol{F}_{l}\in\mathbb{R}^{\frac{H}{4}\times\frac{W}{4}\times4C}$. The latent features will be combined with encoded features for restoration, resulting in the final output $\boldsymbol{F_{out}}\in\mathbb{R}^{\mathrm{H\times W\times C}}$. The overall architecture of D2Net is a 4-level hierarchical encoder-decoder UNet-like~\cite{ronneberger2015u} structure. During the encoding stage, the feature maps progressively reduce the spatial resolution while increasing the channel capacity. Conversely, during the decoding stage, the process is reversed, where the spatial resolution is increased while the channel capacity decreases. Unlike the simple stacking of features used in previous UNet-like approaches, we design the Adaptive Feature Modulation Module to adjust the fusion of encoded and decoded features dynamically. Lastly, the final features $\boldsymbol{F_{out}}$ are refined to obtain the residual image $\boldsymbol{R}$ to which degraded image is added to obtain the normal-light image: $\boldsymbol{y}=\boldsymbol{x}+\boldsymbol{R}$.

\begin{figure}[t!]
    \centering
    \includegraphics[width=\linewidth]{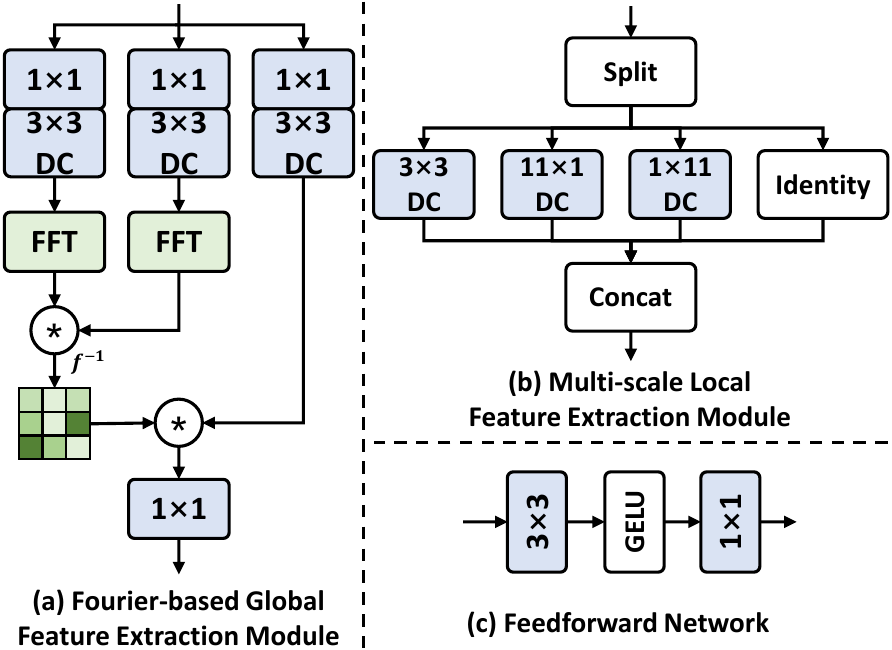}
    \vspace{-18pt}
    \caption{The illustration of (a) Fourier-based Global Feature Extraction (FGFE) Module, (b) Multi-scale Local Feature Extraction (MLFE) Module and (c) Feedforward Network (FFN).}
    \label{fig:module}
\end{figure}

\subsection{Fourier-based Global Feature Extraction Module}
FGFE module aims to capture long-range dependencies of features. For UHD images, modeling long-range dependencies between features is a highly challenging task due to the high-resolution nature of UHD images and this often leads to memory overflow issues. This motivates us to develop an efficient global modeling approach to address the modeling of long-range dependencies between features. We observe that when an image is transformed from the spatial domain to the Fourier domain, the features in the Fourier domain inherently exhibit interactions. Therefore, leveraging the Fourier transform, we design an efficient global modeling module, and the architecture is shown in Fig.~\ref{fig:module} (a). Following the design principles of self-attention~\cite{transformer}, we first map the feature maps through a convolutional group to obtain $\boldsymbol{Q}$, $\boldsymbol{K}$ and $\boldsymbol{V}$. Next, we use Fourier transform $\mathcal{F}(\cdot)$ to convert the spatial domain representation of $\boldsymbol{Q}$ and $\boldsymbol{K}$ into the frequency domain. We compute the frequency attention map $\boldsymbol{M}$ in the Fourier domain and then we apply inverse Fourier transform $\mathcal{F}^{-1}(\cdot)$ to return from the frequency domain to the spatial domain and compute the final result by combining the transformed attention map with $\boldsymbol{V}$. Given the input feature maps $\boldsymbol{F_{in}}$, this process is summarized as:
\begin{align}
&\boldsymbol{Q}=\boldsymbol{F_{in}}\cdot\boldsymbol{W_{Q}},\boldsymbol{K}=\boldsymbol{F_{in}}\cdot\boldsymbol{W_{K}},\boldsymbol{V}=\boldsymbol{F_{in}}\cdot\boldsymbol{W_{V}},\\
&\boldsymbol{F_{out}}=\operatorname{Conv}(\boldsymbol{V}\cdot\mathcal{F}^{-1}(\mathcal{F}(\boldsymbol{Q})\cdot\mathcal{F}(\boldsymbol{k}))),
\end{align}
where $\boldsymbol{W_{Q}}$, $\boldsymbol{W_{K}}$ and $\boldsymbol{W_{V}}$ are different projection matrices. $\operatorname{Conv}(\cdot)$ stands for the $1\times1$ convolution layer. A convolutional group consists of a $1\times1$ depth-wise convolution followed by a $3\times3$ convolution.

\begin{figure*}[t!]
    \centering
    \includegraphics[width=\linewidth]{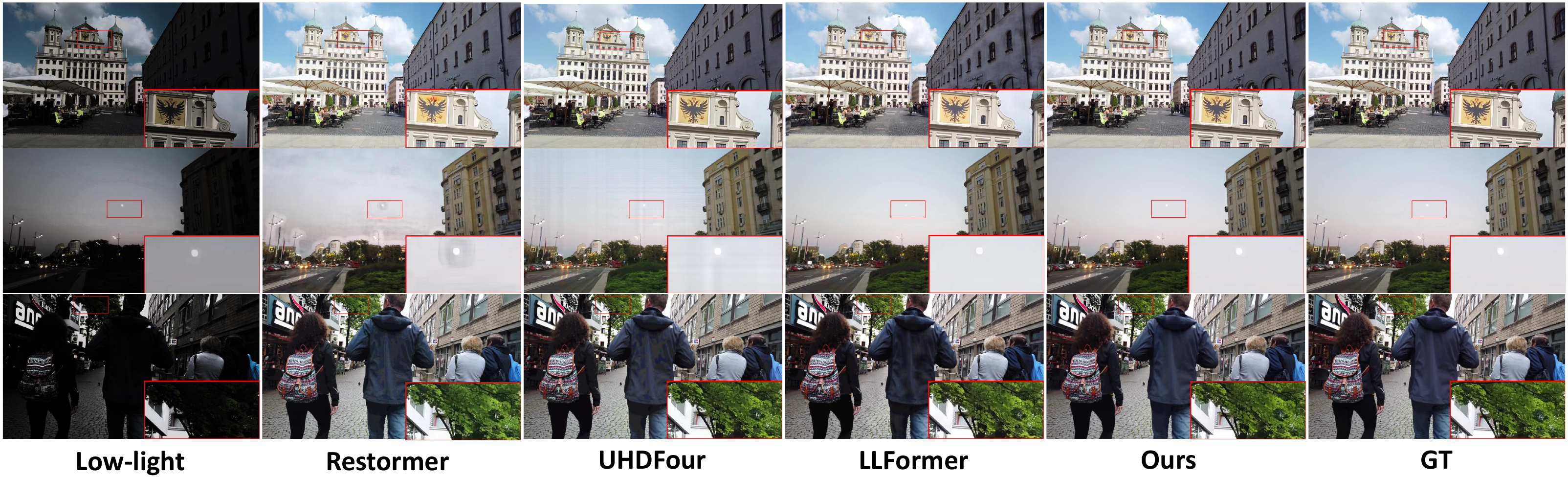}
    \vspace{-18pt}
    \caption{Visual quality comparisons with state-of-the-art methods on UHD-LOL4K dataset. Please zoom in for details.}
    \label{fig:lowlight}
\end{figure*}

\subsection{Multi-scale Local Feature Extraction Module}
Due to the high-resolution nature of UHD images, they contain richer and more detailed patterns compared to images with regular resolutions. In addition, smaller convolution kernel sizes would make it difficult to capture high-quality local features in UHD images, as their high-resolution characteristics require a larger receptive field. Therefore, we design the MLFE module and the architecture is shown in Fig.~\ref{fig:module} (b). To reduce computational complexity, we employ depth-wise convolutions in this module and selectively process only a subset of channels~\cite{yu2023inceptionnext}. For input $\boldsymbol{F_{in}}$, we split it into four groups along the channel dimension,
\begin{align}
\begin{aligned}
    \boldsymbol{F_{hw}}, \boldsymbol{F_{w}}, \boldsymbol{F_{h}}, \boldsymbol{F_{id}}&=\operatorname{Split}(\boldsymbol{F_{in}})\\&= \boldsymbol{F_{in::g}},\boldsymbol{F_{in:g:2g}},\boldsymbol{F_{in:2g:3g}},\boldsymbol{F_{in:3g:}},
\end{aligned}
\end{align}
where $\boldsymbol{g}$ is the channel numbers of convolution branches. We set a ratio $\boldsymbol{r_{g}}$ to determine the branch channel numbers by $\boldsymbol{g}=\boldsymbol{r_{g}}\cdot \boldsymbol{C}$. $\operatorname{Split}(\cdot)$ is the feature splitting operation. Next, the splitting inputs are fed into different parallel branches, this process can be formulated as:
\begin{align}
    \begin{aligned}
    &\boldsymbol{F_{\mathrm{hw}}^{\prime}} =\mathrm{DWConv}_{k_{s}\times k_{s}}^{g\to g}g(\boldsymbol{F_{\mathrm{hw}}}), \\
    &\boldsymbol{F_{\mathrm{w}}^{\prime}}=\mathrm{DWConv}_{1\times k_{b}}^{g\to g}g(\boldsymbol{F_{\mathrm{w}}}), \\
    &\boldsymbol{F_{\mathrm{h}}^{\prime}}=\mathrm{DWConv}_{k_{b}\times1}^{g\to g}g(\boldsymbol{F_{\mathrm{h}}}), \\
    &\boldsymbol{F_{\mathrm{id}}^{\prime}}=\boldsymbol{F_{\mathrm{id}}},
    \end{aligned}
\end{align}
where $\boldsymbol{k_{s}}$ denotes the small square kernel size and $\boldsymbol{k_{b}}$ represents the band kernel size. $\operatorname{DWConv}(\cdot)$ is the depth-wise convolution. Finally, the outputs from each branch are concatenated,
\begin{equation}
    \boldsymbol{F_{out}}=\operatorname{Concat}(\boldsymbol{F_{\mathrm{hw}}^{\prime}}, \boldsymbol{F_{\mathrm{w}}^{\prime}}, \boldsymbol{F_{\mathrm{h}}^{\prime}}, \boldsymbol{F_{\mathrm{id}}^{\prime}}),
\end{equation}
where $\operatorname{Concat}(\cdot)$ denotes the feature concatenating operation.

\subsection{Adaptive Feature Modulation Module}
Previous Unet-like methods typically employ a simple concatenation of encoded and decoded features for processing, which we consider to be a crude approach. In the process of obtaining encoded features at each level in the Unet-like structure, there is a progression from coarse to fine-grained for the feature maps. This means that the feature maps obtained in the initial few steps contain a lot of irrelevant information. Therefore, we design the AFMM to reduce the flow of irrelevant information. The encoded and decoded features at the same level are first aggregated together to generate interactions. Subsequently, a softmax operation is applied to obtain the corresponding weight maps. Based on the weight maps, the encoded and decoded features adaptively adjust their feature maps and are then summed together to form the final feature representations. Given the encoded features $\boldsymbol{F_{en}}$ and decoded features $\boldsymbol{F_{de}}$ at the same level, this process is summarized as:
\begin{align}
    &\boldsymbol{M}=\operatorname{Conv_{2}}(\operatorname{Concat}(\operatorname{Conv_{0}}(\boldsymbol{F_{en}}),\operatorname{Conv_{1}}(\boldsymbol{F_{de}}))),\\
    &\boldsymbol{F_{out}}=\operatorname{Conv_{0}}(\boldsymbol{F_{en}})\cdot\operatorname{S_{0}}(\boldsymbol{M})+\operatorname{Conv_{1}}  \cdot\operatorname{S_{1}}(\boldsymbol{M}),
\end{align}
where $\operatorname{Conv}(\cdot)$ denotes the $1\times1$ convolution and $\operatorname{S}(\cdot)$ is the Softmax operation.

\subsection{Feedforward Network}
FFN is designed to map features into a more compact representation~\cite{MixNet,sun2023spatially}. It consists of a $3\times3$ convolution, a $1\times1$ convolution, and the GULE activation function. FFN initially increases the number of channels in the feature maps by several times. After applying the GULE activation function, the number of channels is then adjusted back to the original size. The entire process can be represented as follows:
\begin{equation}
    \boldsymbol{F_{out}}=\operatorname{Conv_{1\times1}}(\operatorname{GELU}(\operatorname{Conv_{3\times3}\boldsymbol{(F_{in}})})),
\end{equation}
where $\boldsymbol{F_{in}}$ and $\boldsymbol{F_{out}}$ are the input features and output features, respectively.

\begin{figure*}[t!]
    \centering
    \includegraphics[width=\linewidth]{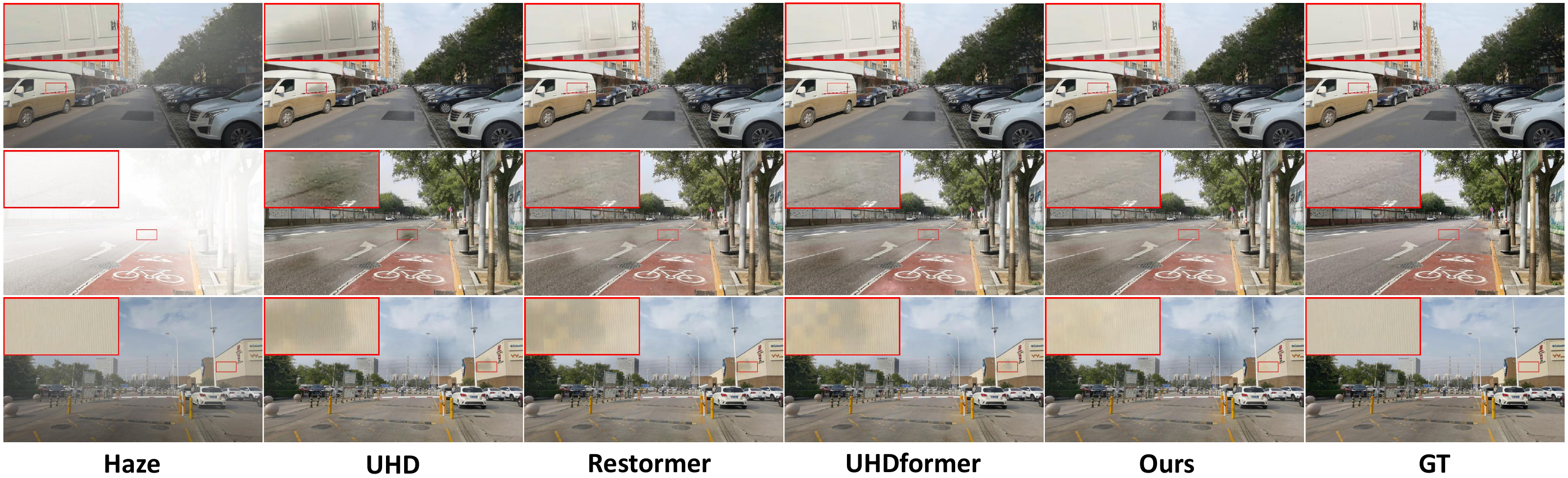}
    \vspace{-18pt}
    \caption{Visual quality comparisons with state-of-the-art methods on UHD-Haze dataset. Please zoom in for details.}
    \label{fig:haze}
\end{figure*}

\begin{table}[t!]
\caption{Comparison of quantitative results on UHD-LOL4K dataset. Best and second best values are indicated with \textbf{bold} text and \underline{underlined} text respectively.}
\vspace{-10pt}
\label{tab:UHDLOL}
\resizebox{\linewidth}{!}{
\begin{tabular}{c|c|c|cc|c}
\toprule[0.5mm]
Methods   & Type                                          & Venue    & PSNR  & SSIM & Param \\ \hline
Z\_DCE++  & \multicolumn{1}{c|}{\multirow{5}{*}{non-UHD}} & TPAMI'21 & 15.58 & 0.934 & 79.42K\\
RUAS      & \multicolumn{1}{c|}{}                         & CVPR'21  & 14.68 & 0.757 & 3.44K\\
ELGAN     & \multicolumn{1}{c|}{}                         & TIP'21   & 18.36 & 0.864 & -\\
Uformer   & \multicolumn{1}{c|}{}                         & CVPR'22  & 29.98 & 0.980 & 20.63M\\
Restormer & \multicolumn{1}{c|}{}                         & CVPR'22  & 36.90 & 0.988 & 26.11M\\ \hline
NSEN      & \multicolumn{1}{c|}{\multirow{4}{*}{UHD}}     & MM'23    & 29.49 & 0.980 & 2.67M\\
UHDFour   & \multicolumn{1}{c|}{}                         & ICLR'23  & 36.12 & \underline{0.990} & 17.54M\\
LLFormer  & \multicolumn{1}{c|}{}                         & AAAI'23  & \underline{37.33} & 0.988 & 24.52M \\
Ours      & \multicolumn{1}{c|}{}                         & -        & \textbf{37.73} & \textbf{0.992} & 5.22M\\
\bottomrule[0.5mm]
\end{tabular}}
\end{table}

\section{Experiment}
\subsection{Implementation Details and Datasets}
\noindent\textbf{Implementation Details.} We conduct experiments in PyTorch on eight NVIDIA GeForce RTX 3090 GPUs. The model is trained with the Adam optimizer~\cite{kingma2014adam} with $\beta_1=0.9$ and $\beta_2=0.999$ for 300k iterations. The learning rate is initially set to $2\times10^{-4}$ and employs the cosine annealing scheme during the training process. We randomly crop the full-resolution 4K image to a resolution of $512\times512$ as the input with a batch size of $16$. To augment the training data, random horizontal and vertical flips are applied to the input images. The architecture of our method consists of a 4-level encoder-decoder, with varying numbers of FEMs at each level, specifically [2, 4, 4, 6] from level-1 to level-4. The refinement stage contains 2 blocks and feature channels set to 24. For the FGFE module, we focus on a patch size of $8\times8$ pixels when computing the frequency attention map. The training objective is to minimize the mean absolute error (MAE, \ie, L1 loss) between the restored image and ground truth.

\noindent\textbf{Datasets.} To verify the effectiveness of D2Net, we evaluate it on the UHD-LOL dataset~\cite{LLFormer}, UHD-Haze dataset and UHD-Blur dataset. The UHD-LOL dataset serves as a comprehensive benchmark for evaluating the performance of D2Net in low-light conditions. It consists of two subsets, namely UHD-LOL4K and UHD-LOL8K, which contain UHD images with resolutions of 4K and 8K, respectively. In this study, we focus on the UHD-LOL4K subset to validate the effectiveness of D2Net. The UHD-LOL4K subset comprises a total of 8,099 image pairs, with 5,999 pairs allocated for training purposes and 2,100 pairs designated for testing. UHD-Haze and UHD-Blur datasets are respectively re-collected from the datasets of Zheng~\etal~\cite{UHDHaze} and Deng~\etal~\cite{deng2021multi} by Wang~\etal~\cite{UHDformer}. Among them, the UHD-Haze dataset has 2,290 haze-clear image pairs for training and 231 pairs for testing. The UHD-Blur dataset has 1,964 blur-clear image pairs for training and 300 pairs for testing.

\subsection{Comparisons with the State-of-the-art Methods}
\noindent\textbf{Evaluation.} 
Two well-known metrics, Peak Signal-to-Noise Ratio (PSNR) and Structural Similarity (SSIM)~\cite{SSIM}, are employed for quantitative comparisons. Higher values of these metrics indicate superior performance of the methods. PSNR and SSIM are calculated along the RGB channels. We also report the trainable parameters (Param). For methods that cannot perform full-resolution inference, we have adopted the previous patch-based or downsample-based approaches to handle them.

\noindent\textbf{Low-Light UHD Image Enhancement Results.} 
We evaluate UHD low-light image enhancement results on UHD-LOK4K dataset with, and compare our method with approaches Z\_DCE++~\cite{ZDCE}, RUAS~\cite{RUAS}, ELGAN~\cite{ELGAN}, Uformer~\cite{Uformer}, Restormer~\cite{Restormer}, NSEN~\cite{NSEN}, UHDFour~\cite{UHDFour} and LLFormer~\cite{LLFormer}. As shown in Tab.~\ref{tab:UHDLOL}, our method outperforms the current state-of-the-art method, LLFormer, by 0.4 dB in PSNR. And you can find the visualization results in Fig.~\ref{fig:lowlight}.

\noindent\textbf{UHD Image Dehazing Results.} 
For the UHD image dehazing task, we compare our method to recent approaches such as Restormer~\cite{Restormer}, Uformer~\cite{Uformer}, DehazeFormer~\cite{dehazeformer}, UHD~\cite{UHDHDR} and UHDformer~\cite{UHDformer}. And the results are shown in Tab.~\ref{tab:UHDHaze}. Compared to the latest method, UHDformer, our approach achieved a 2.29 dB improvement in PSNR, which is undoubtedly a significant improvement. Compared to previous methods, D2Net can better achieve color correction, as you can see from the qualitative comparison results in Fig.~\ref{fig:haze}.

\begin{figure*}[t!]
    \centering
    \includegraphics[width=\linewidth]{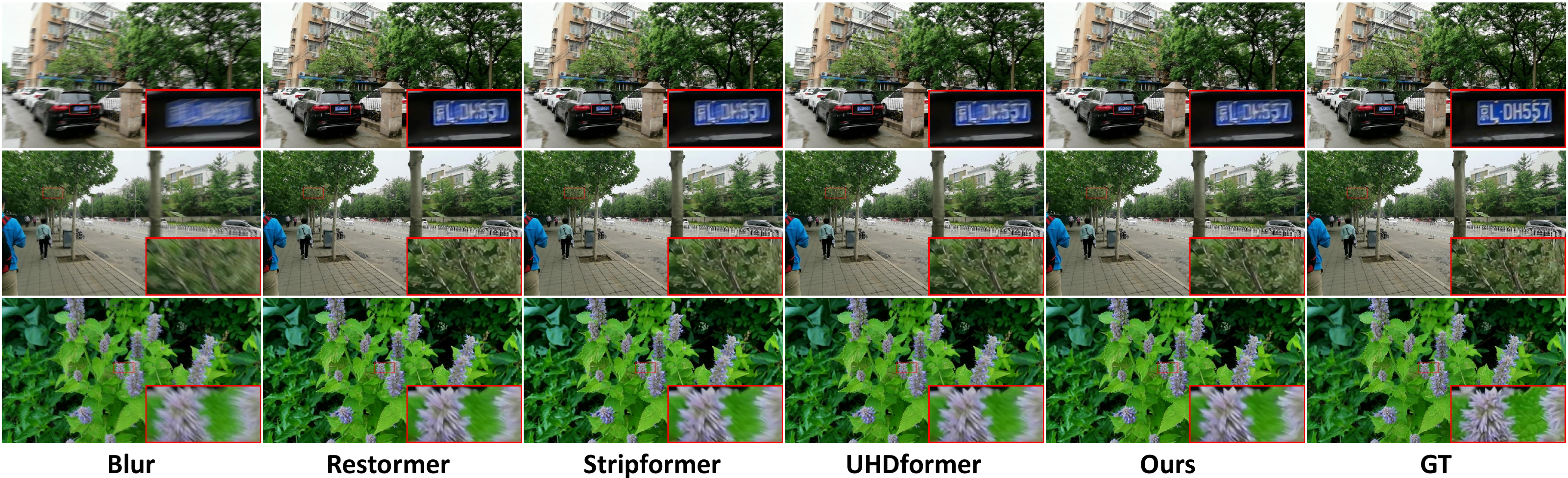}
    \vspace{-18pt}
    \caption{Visual quality comparisons with state-of-the-art methods on UHD-Blur dataset. Please zoom in for details.}
    \label{fig:blur}
\end{figure*}

\begin{table}[t!]
\caption{Comparison of quantitative results on UHD-Haze dataset. Best and second best values are indicated with \textbf{bold} text and \underline{underlined} text respectively.}
\vspace{-10pt}
\label{tab:UHDHaze}
\resizebox{\linewidth}{!}{
\begin{tabular}{c|c|c|cc|c}
\toprule[0.5mm]
Methods   & Type                                          & Venue    & PSNR  & SSIM & Param \\ \hline
Restormer      & \multicolumn{1}{c|}{\multirow{3}{*}{non-UHD}}    & CVPR'22  & 12.72 & 0.693 & 26.11M\\
Uformer     & \multicolumn{1}{c|}{}                         & CVPR'22   & 19.83 & 0.737 & 20.63M\\
DehazeFormer   & \multicolumn{1}{c|}{}                         & TIP'23  & 15.37 & 0.725 & 2.5M\\ \hline
UHD      & \multicolumn{1}{c|}{\multirow{3}{*}{UHD}}     & ICCV'21    & 18.04 & 0.811 & 34.5M\\
UHDformer      & \multicolumn{1}{c|}{}     & AAAI'24    & \underline{22.59} & \underline{0.943} & 0.34M \\
Ours      & \multicolumn{1}{c|}{}                         & -        & \textbf{24.88} & \textbf{0.944} & 5.22M\\
\bottomrule[0.5mm]
\end{tabular}}
\end{table}

\noindent\textbf{UHD Image Deblurring Results.} 
In the UHD image deblurring task, we evaluate our proposed D2Net against existing deblurring approaches, including MIMO-Unet++~\cite{MIMO}, Restormer~\cite{Restormer}, Uformer~\cite{Uformer}, Stripformer~\cite{Stripformer}, FFTformer~\cite{fftformer} and UHDformer~\cite{UHDformer}. As shown in Tab.~\ref{tab:UHDBlur}, our method achieved a 1.64 dB improvement in PSNR compared to the current state-of-the-art method, UHDformer. Additionally, you can find the visual comparison results in Fig.~\ref{fig:blur}.

\begin{figure*}[t!]
    \centering
    \includegraphics[width=.9\linewidth]{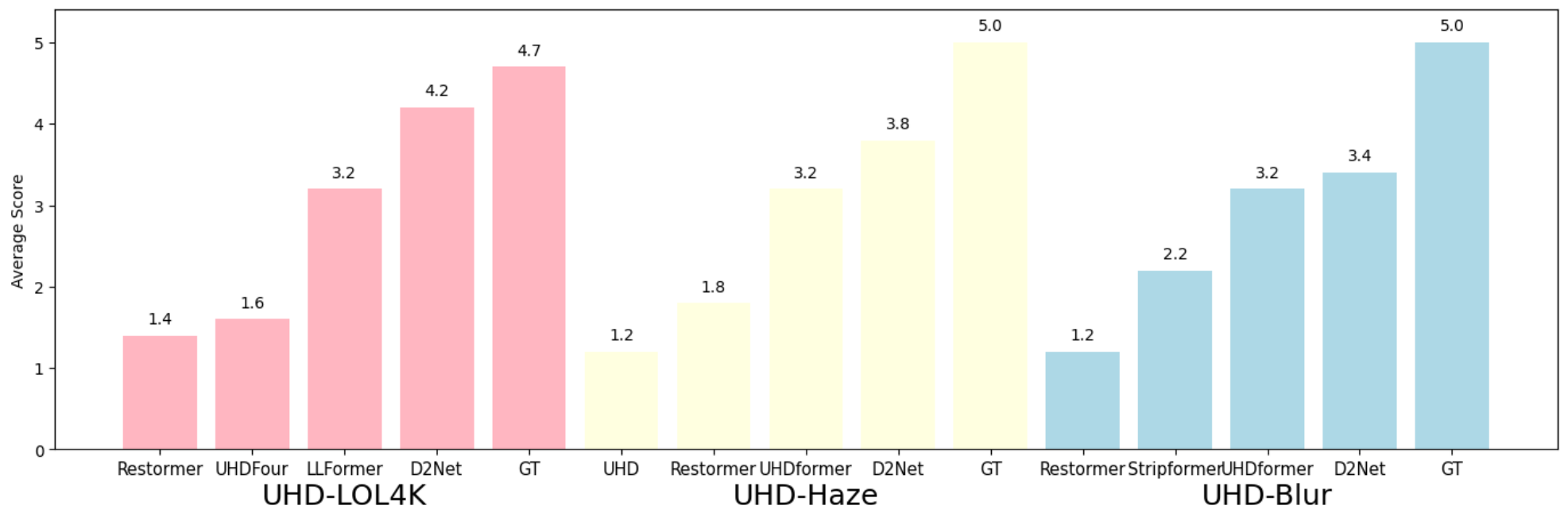}
    \vspace{-14pt}
    \caption{User study scores on three benchmarks.}
    \label{fig:user}
\end{figure*}

\noindent\textbf{User Study Score.} We conduct a user study to quantify the human subjective visual perception quality of the restored UHD images from the three datasets. 5 human subjects are invited to score the visual quality of 15 restored images, independently. Specifically, we allocate 5 restored images per task, which consist of the results of 4 different methods and the ground truth. These subjects are instructed to subjectively rank the quality of these 5 images, and The scores range from 1 (worst) to 5 (best). From Fig.~\ref{fig:user}, we can see the results of user study. From the results, we can draw two conclusions: 1) Our approach has a more realistic and precise visual quality compared to other methods. 2) The gap between the current methods' restored images and the GT is still relatively large, and further improvement is needed.

\begin{table}[t!]
\caption{Comparison of quantitative results on UHD-Blur dataset. Best and second best values are indicated with \textbf{bold} text and \underline{underlined} text respectively.}
\vspace{-10pt}
\label{tab:UHDBlur}
\resizebox{\linewidth}{!}{
\begin{tabular}{c|c|c|cc|c}
\toprule[0.5mm]
Methods   & Type                                          & Venue    & PSNR  & SSIM & Param \\ \hline
MIMO-Unet++  & \multicolumn{1}{c|}{\multirow{5}{*}{non-UHD}} & ICCV'21 & 25.03 & 0.752 & 16.1M\\
Restormer      & \multicolumn{1}{c|}{}                         & CVPR'22  & 25.21 & 0.752 & 26.1M\\
Uformer     & \multicolumn{1}{c|}{}                         & CVPR'22   & 25.27 & 0.752 & 20.6M\\
Stripformer   & \multicolumn{1}{c|}{}                         & ECCV'22  & 25.05 & 0.750 & 19.7M\\
FFTformer & \multicolumn{1}{c|}{}                         & CVPR'23  & 25.41 & 0.757 & 16.6M\\ \hline
UHDformer      & \multicolumn{1}{c|}{\multirow{2}{*}{UHD}}     & AAAI'24    & \underline{28.82} & \underline{0.844} & 0.34M\\
Ours      & \multicolumn{1}{c|}{}                         & -        & \textbf{30.46} & \textbf{0.872} & 5.22M\\
\bottomrule[0.5mm]
\end{tabular}}
\end{table}

\subsection{Ablation Study}
We further conduct extensive ablation studies to understand better and evaluate each component in the proposed D2Net. To ensure a fair comparison, all experiments are conducted on the UHD-LOL4K dataset. In order to demonstrate the effectiveness of the proposed module, we replace it with the Residual Block (ResBlock)~\cite{he2016deep} of comparable parameters.

\noindent\textbf{Effectiveness of Multi-scale Local Feature Extraction Modules.} To demonstrate the gains of MLFE modules, we repalce it with the ResBlock of comparable parameters. The results shown in Tab.~\ref{table:abla} demonstrate the effectiveness of the MLFE modules. Multi-scale feature representation to provide performance gains is already a common practice~\cite{yu2023inceptionnext,hui2019lightweight}. However, existing UHD image restoration methods have overlooked this aspect. For UHD images, it is even more important, as at higher resolutions, convolutions with different receptive fields can more easily capture different patterns.

\noindent\textbf{Effectiveness of Fourier-based Global Feature Extraction Modules.} To demonstrate the gains of FGFE modules, we repalce it with the ResBlock of comparable parameters. From the results in Tab.~\ref{table:abla}, we can see that the FGFE modules bring significant performance gains, which are greater than the gains brought by the MLFE modules. This suggests that the performance gains from non-local features are substantial, which is likely why researchers have been trying to apply self-attention to various tasks. Furthermore, we will discuss the relationship between self-attention and our FGFE modules.

Both self-attention and the FGFE modules aim to establish long-range feature dependencies. So why do we not try using self-attention and instead take a different approach by modeling global features in the frequency domain? LLFormer~\cite{LLFormer} has already tried using self-attention to solve the UHD image restoration problem, but encountered issues with memory overflow, forcing them to divide the image into patches and process them individually. The main factor causing the memory overflow is the storage of the attention map. For self-attention, the spatial complexity is generally $\textit{O}(HW\times HW)$. And our FGFE modules only require $\textit{O}(HW)$, which is sufficient to model the long-range feature dependencies, thereby successfully avoiding the memory overflow issue. There are also methods that attempt to calculate the attention map along the channel dimension to reduce memory consumption~\cite{Restormer}. However, these methods often require stacking multiple attention blocks to achieve performance comparable to the methods that compute attention maps along the spatial dimension. This approach is also not suitable for UHD images.

\noindent\textbf{Effectiveness of Adaptive Feature Modulation Modules.} 
We compare our proposed AFMM with the commonly used direct feature concatenation approach in the UNet structure. The results are shown in Tab.~\ref{table:abla2}. Based on the observed results, our proposed AFMM is clearly superior to the simple approach of stacking features together. The effectiveness of AFMM, in our view, may be attributed to the inherent redundancy in image data, where the same patterns can appear multiple times in a single image. The encoded features may contain some irrelevant information, and the flow of such information can degrade the model's performance. Therefore, the adaptive selection of features in AFMM proves to be effective in mitigating this issue. 

\begin{figure}[t!]
    \centering
    \includegraphics[width=\linewidth]{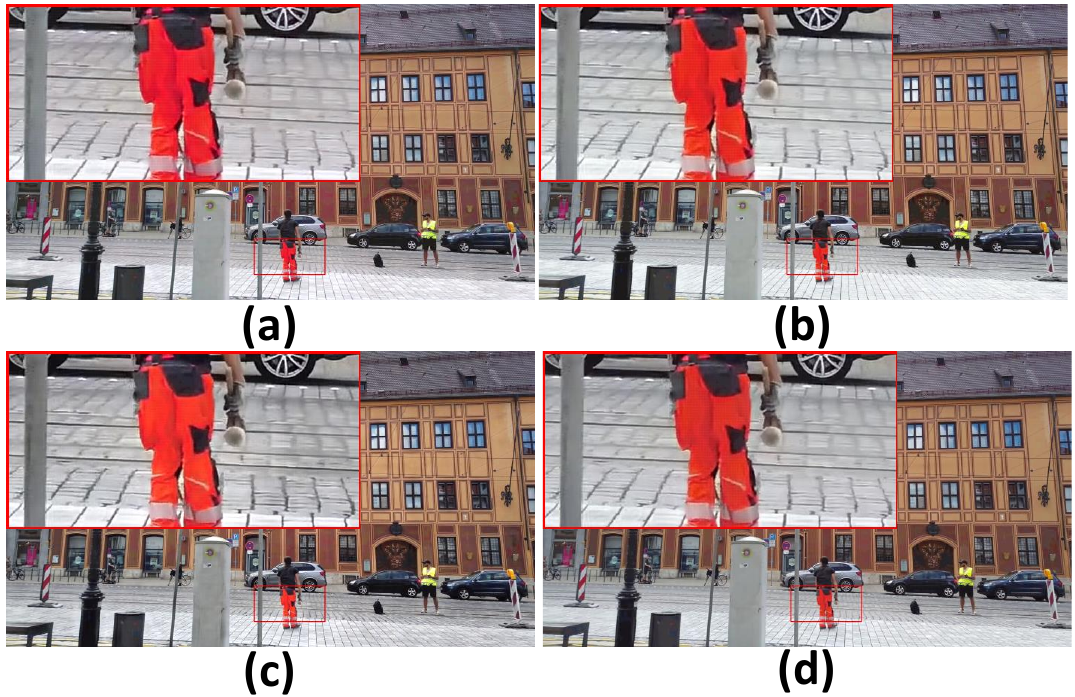}
    \vspace{-20pt}
    \caption{Visual quality comparisons of ablation study.  Please zoom in for details. The labels in the figure correspond to the labels in Tab.~\ref{table:abla}.}
    \label{fig:abla}
\end{figure}

\begin{table}[t!]
\centering
\caption{Ablation study of proposed blocks. In order to ensure consistency in parameter magnitude, we gradually replace the proposed module with Residual Block of comparable parameters.}
\vspace{-10pt}
\resizebox{.7\linewidth}{!}{
\label{table:abla}
\begin{tabular}{c|cc|cc}
\toprule[0.5mm]
\#  & FGFE       & MLFE       & PSNR  & SSIM  \\ \hline
(a) &            &            & 36.88 & 0.984 \\
(b) & \checkmark &            & 37.47 & 0.988 \\
(c) &            & \checkmark & 37.13 & 0.984 \\
(d) & \checkmark & \checkmark & 37.73 & 0.992 \\
\bottomrule[0.5mm]
\end{tabular}}
\end{table}
\begin{table}[t!]
\centering
\vspace{-10pt}
\caption{Ablation study of Adaptive Feature Modulation Module.}
\vspace{-10pt}
\resizebox{.5\linewidth}{!}{
\label{table:abla2}
\begin{tabular}{c|cc}
\toprule[0.5mm]
       & PSNR  & SSIM  \\ \hline
Concat & 36.97 & 0.989 \\
AFMM   & 37.73 & 0.992 \\
\bottomrule[0.5mm]
\end{tabular}}
\vspace{-10pt}
\end{table}
\begin{table}[t!]
\centering
\caption{Trainable parameters,running time and inference memory cost of some UHD image restoration methods.}
\vspace{-10pt}
\label{tab:time}
\resizebox{.8\linewidth}{!}{
\begin{tabular}{c|cccc}
\toprule[0.5mm]
      & UHD   & UHDFour & UHDformer & Ours  \\ \hline
Param & 34.5M & 17.54M  & 0.34M     & 5.22M \\
RT    & 1.43s & 1.67s   & 5.90s     & 1.63s \\ 
Mem   & 4.50G & 2.02G   & 11.46G    & 10.97G \\\toprule[0.5mm]
\end{tabular}}
\end{table}

\section{Conclusion}
In this paper, we proposed a novel UHD image restoration framework, named D2Net. D2Net enables direct full-resolution inference on consumer-grade GPUs, without the need for high-rate downsampling or dividing the images into several patches. Specifically, we ingeniously leveraged the characteristics of the Fourier transform to achieve long-range dependency modeling of features in the frequency domain. We also designed an MLFE module to extract multi-scale local features from the images. Additionally, we introduced the AFMM to reduce the flow of irrelevant information in the Unet-like structure. Extensive experiments showed the superiority of D2Net on several UHD image restoration tasks.

\noindent\textbf{Limitations.} 
Full-resolution inference can improve the quality of the restored images, but it also results in a significant memory overhead. In Tab.~\ref{tab:time}, we report the training parameters, inference time, and memory overhead required for recent UHD image restoration methods to process $3840\times2160$ resolution images on a single NVIDIA GeForce RTX 3060 GPU. From the results, we can see that our method, like most previous approaches, incurs a relatively large inference memory cost when processing a single UHD image. Looking ahead, we will focus on optimizing the inference memory overhead of our method.

{\small
\bibliographystyle{ieee_fullname}
\bibliography{egbib}
}

\end{document}